\documentclass{Interspeech2024}
\usepackage[utf8]{inputenc} 
\usepackage[T1]{fontenc}    
\usepackage{url}            
\usepackage{booktabs}       
\usepackage{amsfonts}       
\usepackage{nicefrac}       
\usepackage{microtype}      
\usepackage{xcolor}         
\usepackage{multirow}
\usepackage{amsmath}
\usepackage{mathtools}
\usepackage{graphicx}
\usepackage{caption}
\usepackage{subcaption}
\usepackage{pgfplots}
\usepackage{url}
\usepackage{tikz}
\usepackage{todonotes}
\usepackage{etoolbox}

\interspeechcameraready

\title{Linear-Complexity Self-Supervised Learning for Speech Processing}

\name[affiliation={}]{Shucong}{Zhang}
\name[affiliation={}]{Titouan}{Parcollet}
\name[affiliation={}]{Rogier}{van Dalen}
\name[affiliation={}]{Sourav}{Bhattacharya}
\address{
    Samsung AI Center Cambridge, United Kingdom}

\email{\{s1.zhang|t.parcollet|r.vandalen|sourav.b1\}@samsung.com}


\newcommand{\Reals}{\mathbb R}

\newcommand{\Output}{\mathbf H}
\newcommand{\OutputVector}{\mathbf h}
\newcommand{\Input}{\mathbf X}
\newcommand{\InputVector}{\mathbf x}

\newcommand{\SummaryMean}{\bar{\mathbf s}}

\newcommand{\Transformation}{f}

\newcommand{\Summary}{s}

\newcommand{\CombineFunction}{c}

\newcommand{\Time}{t}

\newcommand{\Length}{T}

\newcommand{\InputDimensionality}{D}
\newcommand{\OutputDimensionality}{D'}
\newcommand{\TransformedDimensionality}{D''}

\begin{document}

\maketitle

\begin{abstract}
Self-supervised learning (SSL) models usually require weeks of pre-training with dozens of high-end GPUs. These models typically have a multi-headed self-attention (MHSA) context encoder. However, MHSA takes quadratic time and space in the input length, contributing to the high pre-training cost. Linear-complexity alternatives to MHSA have been proposed. For instance, in supervised training, the SummaryMixing model is the first to outperform MHSA across multiple speech processing tasks. However, these cheaper alternatives have not been explored for SSL yet. This paper studies a linear-complexity context encoder for SSL for the first time. With better or equivalent performance for the downstream tasks of the MP3S benchmark, SummaryMixing reduces the pre-training time and peak VRAM of wav2vec 2.0 model by 18\% and  by 23\%, respectively, leading to the pre-training of a 155M wav2vec 2.0 model finished within one week with 4 Tesla A100 GPUs. Code\footnote{\url{https://github.com/SamsungLabs/SummaryMixing}.} is available.
\end{abstract}

\section{Introduction}

Self-supervised learning (SSL) models have demonstrated state-of-the-art (SOTA) performance for speech processing tasks \cite{baevski2020wav2vec, hsu2021hubert, chiu2022self,  baevski2023efficient}. 
SSL models are pre-trained on unlabeled data to learn hidden features of the input audio. 
Since the pre-training does not require human transcription, SSL can leverage a huge amount of unlabeled data.
The large amount of pre-training data is one of the keys to the success of SOTA SSL models. 
SSL models also benefit from the large model size.
The typical model size varies from millions to billions parameters. 

However, training large models with huge amount of data leads to extremely big training cost,
resulting in an exceedingly high barrier for the research of SSL models, as well as extensive carbon footprints.
For example, the pre-training of a 330M wav2vec 2.0 model \cite{baevski2020wav2vec} with 3k hours data requires 32 Tesla V100 GPUs running for two weeks, consuming 1.818 MWh of energy,
while the pre-training of a 965M wav2vec 2.0 model with 14k hours data requires 32 Tesla A100 GPUs running for two weeks, consuming 16.511 MWh of energy \cite{parcollet2024lebenchmark}.

In this paper, we address the pre-training inefficiency of SSL models from the efficient architecture aspect.
SOTA SSL models typically consistent of a feature extractor and a context encoder.
The feature extractor extracts features from the raw wave input, and the context encoder generates further hidden representations.
Efficient feature extractors have been proposed \cite{chiu2022self, barrault2023seamlessm4t, parcollet2023efficiency, lin2023melhubert}.
Nevertheless, to the best of our knowledge, the context encoder has not been studied from the efficiency angle yet. 
Thus, this paper improves the pre-training efficiency from the aspect of context encoder.

The context encoder of the SOTA SSL models is usually a multi-headed self-attention (MHSA) Transformer \cite{vaswani2017attention} or Conformer \cite{gulati2020conformer} encoder. 
However, MHSA has a quadratic time and space complexity in the input sequence length, slowing down pre-training and increasing VRAM consumption.
Methods of developing sub-quadratic complexity alternatives to MHSA include designing priors for the attention patterns \cite{zaheer2020big, beltagy2020longformer}, low-rank approximation \cite{wang2020self}, kernelization \cite{katharopoulos2020transformers}, and linearization \cite{wu2021fastformer}.
Unfortunately, compared to MHSA, these methods usually lead to inferior results for speech processing tasks \cite{peng2022branchformer, parcollet2024summarymixing}.
Aggressive downsampling is also commonly used to reduce the training and inference time and VRAM consumption of MHSA based speech processing models \cite{burchi2021efficient, kim2022squeezeformer, rekesh2023fast}. 
Nevertheless, this approach does not reduce the quadratic complexity of MHSA.

However, a recently-developed linear-complexity model, SummaryMixing \cite{parcollet2024summarymixing}, is promising for developing linear-complexity SSL models, since it is the first linear-comlexity model that  surpasses SOTA MHSA models for automatic speech recognition and spoken language understanding under supervised training.
SummaryMixing has two branches: a local branch uses a point-wise feed-forward network to capture the local information, and a summary branch which uses the average vector of the input frames to capture the global information. 
The output of the two branches are merged to form the hidden representations of the input.
Although this efficient model performs well in supervised learning,
it is unknown yet whether this simple design is flexible enough to capture all necessary features at different levels for different downstream speech processing tasks.

In this paper, we equip wav2vec 2.0 using a Conformer context encoder with SummaryMixing.
We show that compared to MHSA Conformer wav2vec 2.0 model, our proposed model gives better or equivalent results for the downstream automatic speech recognition, intent classification, emotion recognition, and automatic speaker verification tasks of the MP3S benchmark \cite{zaiem2023speech}. 
The numerical experimental results and our analysis of downstream tasks demonstrate that SummaryMixing captures different levels of speech representations (i.e.\ content, semantic, paralinguistics, and speaker features) well through SSL pre-training.
For the efficiency aspect, SummaryMixing reduces the pre-training time and the peak VRAM by 18\% and 23\%, respectively, making the pre-training of a 155M wav2vec 2.0 model finished within 7 days with 4 Tesla A100 GPUs.

To the best of our knowledge, this paper is the first to present a linear-complexity SSL model with no performance drop on downstream tasks.
We release the necessary recipes of reproducing SummaryMixing Conformer wav2vec 2.0 with the opensource toolkit SpeechBrain \cite{speechbrain}.

\subsection{Previous works for efficient SSL}
Previous works have addressed the inefficiency of the pre-training of SSL models from different angles.
\cite{chiu2022self, chen2023reducing} improve the pre-training procedure.
\cite{baevski2022data2vec, baevski2023efficient} enhance the efficiency by crafting the pre-training objective.
From the efficient model architecture aspect, the feature extractor have been studied.
\cite{SEW} reduces the number of channels and the kernel size of the deep CNN feature extractor.
A more effective approach is to replace the deep CNN feature extractor with Mel filterbanks or a combination of Mel filterbanks and a shallow CNN \cite{ W2v-bert, chiu2022self, barrault2023seamlessm4t, parcollet2023efficiency, lin2023melhubert}.
Our proposed SummaryMixing context encoder addresses the inefficiency problem from a different angle, and it is compatible with existing methods.
For example, Section~\ref{section: SummaryMixing wav2vec 2.0} will combine an efficient feature extractor with the SummaryMixing context encoder.



\section{SummaryMixing for wav2vec 2.0}
\label{section: SummaryMixing wav2vec 2.0}

This section first introduces SummaryMixing  \cite{parcollet2024summarymixing}. Then, it proposes SummaryMixing wav2vec 2.0 model.

\subsection{SummaryMixing}
\begin{figure}[!htb]
    \begin{subfigure}[b]{.4\columnwidth}
        \centering
        \includegraphics{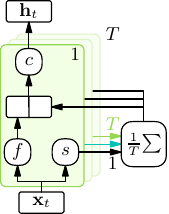}
        \caption{The SummaryMixing block, which can be inserted into the Conformer in Figure~\ref{figure: conformer}.
        The function $\frac1{\Length} \sum$ is executed only once and the average is fed back to each time step $\Time$, leading to a linear complexity.}
        \label{figure: summarymixing}
    \end{subfigure}
    \hspace{\stretch{1}}
    \begin{subfigure}[b]{.55\columnwidth}
        \centering
        \includegraphics{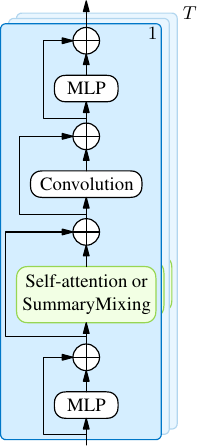}
        \caption{The Conformer, which can be equipped (the green block) with either self-attention or SummaryMixing.}
        \label{figure: conformer}
    \end{subfigure}
    \caption{Architectures of SummaryMixing and the Conformer.}
    \label{fig:architectures}
\end{figure}

SummaryMixing is a linear-complexity alternative to MHSA. 
It transforms the input sequence $\Input \in \Reals^{\Length \times \InputDimensionality} = \{\InputVector_0,\ldots,\InputVector_{\Length}\}$ of $\Length$ feature vectors $\InputVector_{\Time}$ of length $\InputDimensionality$ to a sequence of hidden representations  $\Output \in \Reals^{\Length \times \OutputDimensionality} = \{\OutputVector_0,\ldots,\OutputVector_{\Length}\}$.
It has been found that MHSA in trained Transformer and Conformer layers can behave as feed-forward networks \cite{zhang2021usefulness, shim2022understanding}.
For Branchformer which has one convolutional branch and 
one MHSA branch, the MHSA tends to merely implement an average \cite{peng2022branchformer}.
SummaryMixing therefore explicitly computes an average, but at linear cost.
As shown in Figure~\ref{figure: summarymixing}, it has a branch that generates a single vector summarizing the global information of the input sequence by averaging the non-linear transformation $\Summary(\InputVector_{\Time})$ of each input vector over time ($\frac1{\Length}\sum$).
The local branch uses a non-linear transformation $\Transformation(\cdot)$ to extract the local information for each input vector $\InputVector_{\Time}$.
Then, the single global vector and the local vector $\Transformation(\InputVector_{\Time})$ are combined through a non-linear combination function $\CombineFunction(\cdot)$ to produce the hidden representation $\OutputVector_{\Time}$ for each time step. 
Figure~\ref{figure: summarymixing} illustrates the architecture of SummaryMixing. Mathematically, it can be described as
\begin{align}
    \SummaryMean &= \frac1{\Length} \sum_{\Time=1}^{\Length} \Summary(\InputVector_{\Time})
    ;&
    \OutputVector_{\Time} &= \CombineFunction([\Transformation(\InputVector_{\Time}), \SummaryMean]).
    \label{eq:summarymixing}
\end{align}
where $\Summary: \Reals^{\InputDimensionality} \rightarrow \Reals^{\TransformedDimensionality}$, $\Transformation: \Reals^{\InputDimensionality} \rightarrow \Reals^{\TransformedDimensionality}$, and $\CombineFunction: \Reals^{2\TransformedDimensionality} \rightarrow \Reals^{\OutputDimensionality}$.
$\Summary(\cdot)$, $\Transformation(\cdot)$, and $\CombineFunction(\cdot)$ are neural networks with one hidden layer.
The averaging $\frac1{\Length}\sum$ and the non-liner functions $\Summary(\cdot)$, $\Transformation(\cdot)$, and $\CombineFunction(\cdot)$ have a linear time and space complexity with respect to the input sequence length, making SummaryMixing a linear-complexity model.

SummaryMixing can be used inside a Conformer or a  Branchformer \cite{parcollet2024summarymixing}. 
Figure~\ref{figure: conformer} shows the architecture of the Conformer.
The standard self-attention block inside it can simply be replaced with SummaryMixing.
Both Conformer and Branchformer can be equipped with SummaryMixing and reach SOTA performance for automatic speech recognition and spoken language understanding tasks \cite{parcollet2024summarymixing}.


\subsection{SummaryMixing for wav2vec 2.0}
We use wav2vec 2.0 (w2v2) \cite{baevski2020wav2vec} as the SSL model in this paper, since it is a well-established model and popular for SSL research.
Since w2v2 is implemented in widely used toolkits such as SpeechBrain \cite{speechbrain} and FairSeq \cite{ott2019fairseq}, results are straightforward to reproduce.
Compared to more recently proposed SSL models, w2v2 is still competitive for downstream speech processing tasks \cite{zaiem2023speech}. Section~\ref{sec: exp} will show that our SummaryMixing w2v2 model, which will be proposed in this section, outperforms HuBERT \cite{hsu2021hubert} and data2vec \cite{baevski2022data2vec} for a variety of tasks.

To improve the pre-training efficiency of w2v2 model, we equip the context encoder of w2v2 with SummaryMixing. 
SummaryMixing has been used inside both of Branchformers and Conformers for supervised training. However, the Conformer is more typically used for SSL \cite{W2v-bert, barrault2023seamlessm4t, conformerSSL}.
Thus, we will use the Conformer with   SummaryMixing rather than the Branchformer with SummaryMixing as the context encoder. 
In addition, we do not consider the Transformer context encoder in the original w2v2 model \cite{baevski2020wav2vec},
since Conformer-based architectures have demonstrated better performance for both 
 of SSL models \cite{W2v-bert, barrault2023seamlessm4t, conformerSSL} and supervised trained SummaryMixing models \cite{parcollet2024summarymixing}. 

To make the pre-training more efficient, we replace the original w2v2 deep CNN feature extractor with a combination of Mel filterbanks and a shallow 1D CNN as in \cite{parcollet2023efficiency}. This replacement leads to equivalent performance for downstream tasks in the previous work \cite{parcollet2023efficiency}. 
Other model components and the training objective follows the original w2v2 model \cite{baevski2020wav2vec}.

\section{Experiments}
\label{sec: exp}
As discussed in Section~\ref{section: SummaryMixing wav2vec 2.0}, we build a w2v2 model with a Conformer context encoder. 
For simplicity, we use ``MHSA w2v2'' and ``SummaryMixing w2v2'' to denote w2v2 model with MHSA and SummaryMixing Conformer context encoder, respectively.

\subsection{SSL pre-training}
\label{subsection: SSL pre-training}

\noindent\textbf{Model Architecture and training objective.}
For both of MHSA w2v2 and SummaryMixing w2v2, the Conformer context encoder has 12 layers.
Each Conformer layer has a hidden dimension of 768 for the MHSA or SummaryMixing block, and a hidden dimension of 3072 for the MLP block. 
The kernel size and the stride is 31 and 1 for the Convolution block, respectively.
Each MHSA or SummaryMixing block has 8 heads (multi-headed SummaryMixing is defined in \cite{parcollet2024summarymixing}). 
To further enhance the pre-training efficiency, following \cite{parcollet2023efficiency}, we use the Mel filterbanks with a 1D CNN (FBank-CNN1D) as the feature extractor for the w2v2 models in this paper. 
The FBank-CNN1D feature extractor consist of 80 channel filterbank with 25ms windows and a hop length of 10ms, and a 512-channel two-layered 1D CNN with kernel sizes and strides of (3, 3), (2, 1) respectively.
All other components and the training objective follow the original w2v2 \cite{baevski2020wav2vec}. All models are implemented and pre-trained with SpeechBrain V1.0 \cite{speechbrain}. \\

\noindent\textbf{Pre-training details.} We use the Libri-Light Medium subset as the pre-training data \cite{librilight}.
We use the voice activation detection tool in \cite{librilight} to clip the audio recordings. 
For each audio recording, we discard clips longer than 60s.
Then, we concatenate adjacent clips to sequences with a maximum 30s length.
If a clip is between 30s and 60s, it will not be concatenated to any other clip. 
For example, if a recording is clipped into 10s, 10s, and 35s clips, this recording will give a 20s long and a 35s long training sequence.
In this way, the total amount pre-training data is 4.3k hours. 
We use 4 Tesla A100 GPUs for pre-training.
Each GPU has a batch size 360s.
A gradient accumulation of 4 is used, resulting in a 1.6h total batch size. 
The pre-training takes 300k steps.
We use the Noam \cite{vaswani2017attention} learning rate scheduler with 30k warmup steps and a peak learning rate $5\cdot 10^{-4}$.
All other setups follow the SpeechBrain V1.0 w2v2 pre-training configuration. \\

\noindent\textbf{Pre-training Efficiency.}
Table~\ref{tab:efficiency} shows the pre-training time and peak VRAM for both of the MHSA and SummaryMixing w2v2 models. 
Compared to MHSA, SummaryMixing reduces the pre-training time by 18\%, relatively. 
It worth noting that the pre-training of the 155M SummaryMixing w2v2 model can be finished within 7 days with 4 Tesla A100 GPUs. 
With the same infrastructure, batch size, and number of steps, the training time of the original 95M w2v2 base model is more than 10 days \cite{parcollet2023efficiency}. 
For the memory efficiency, compared to MHSA, SummaryMixing also reduces the peak VRAM by 23\%, relatively. 

Moreover, since each SummaryMixing block is more efficient compared to each MHSA block for each Conformer layer, SummaryMixing will lead to larger efficiency gain when scaling up the models.
We demonstrate this by efficiency experiments of doubling the number of Conformer context encoder layers, leading to a 308M SummaryMixing w2v2  model and a 329M MHSA w2v2 model.

Same as the 12-layered 155M SummaryMixing and 165M MHSA w2v2 models, the 24-layered 308M SummaryMixing model can be trained with a 360s batch size on each GPU.
As Table~\ref{tab:efficiency} shows, doubling the layers of SummaryMixing context encoder increases the peak VRAM from 51GB to 76GB.
The peak VRAM of the 24-layered 308M SummaryMixing model is almost the same as the 12-layered 165 MHSA model.
Training the 329M MHSA w2v2 with a 360s batch size per GPU will lead to out of memory issues.
To fit the 329M MHSA w2v2 within the 80G VRAM limitation of a Tesla A100 GPU, the batch size per GPU needs to be reduced from 360s to at most 250s.

We estimate the pre-training time of the scaled-up models by pre-training each model for one epoch. 
Table~\ref{tab:efficiency} shows the scaling-up will increase the pre-training time of SummaryMixing w2v2 from 7 days to 15 days, and will increase the pre-training time of MHSA w2v2 from 9 days to more than 24 days.
Thus, for the scaled up models, the relative pre-training time reduction from SummaryMixing encoder is increased from 18\% to 35\%.
We do not fully pre-train 24-layered w2v2 models in this paper since the efficiency benefit is already demonstrated even with the smaller models,
and the pre-training of the 329M MHSA w2v2 is too time-consuming for our infrastructure.
\begin{table}[t]
    \caption{Efficiency analysis of wav2vec 2.0 models. GPU hours is the pre-training time with a 1.6h batch size for 300k updates with 4 Tesla A100 GPUs. *~indicates the data is estimated. VRAM is measured with a 360s batch size on a single GPU. OOM denotes ``out of memory''. }
    \label{tab:efficiency}
    \centering
    \begin{tabular}{lccc}
        \toprule
        \textbf{wav2vec 2.0}
        & \textbf{Model size} & \textbf{GPU} & \textbf{VRAM}  \\
        \textbf{Context encoder} & \textbf{millions}
        & \textbf{hours $\downarrow$} & \textbf{ GB $\downarrow$}  \\
        \cmidrule(r){1-2} \cmidrule(l){3-4}
        MHSA & 165  & 207  & 76    \\
        SummaryMixing & 155 & 164 & 54    \\
        \cmidrule(r){1-2} \cmidrule(l){3-4}
        MHSA & 329         & 579\rlap{*}                    & OOM   \\
        SummaryMixing & 308 & 376\rlap{*} & 76    \\
        \bottomrule           
    \end{tabular}
\end{table}

\begin{table*}[!t]
    \centering
    \caption{ The speech recognition word error rates (WER), the accuracy (Acc.) for Intent Classification (IC), Emotion Recognition (ER), and the equal error rate (EER) for Automatic Speaker Verification (ASV) of wav2vec 2.0 models. ``D.c'',``T.c'' and ``T.o'' stand for ``dev-clean'',``test-clean'' and ``test-other'' respectively. ASR results are from greedy decoding without language model fusion. The MHSA and SummaryMixing wav2vec 2.0 models are trained and evaluated, while the performance of all other models is from litterateur \cite{zaiem2023speech}. ``LL'', ``LS'', and ``LV'' denote  ``Libri-Light'', ``LibriSpeech'' and ``LibriVox'', respectively.
    For our pre-training experiments, as shown in Section~\ref{subsection: SSL pre-training}, we pre-process the LL medium (5.2k h) subset and this results in 4.3k h pre-training data.}
    \label{tab:res_asr}
    \begin{tabular}{@{~~}l@{~~}c@{~~}l @{~~~~} c@{~~}c@{~~}c @{~~~~} c@{~~}c@{~~}c @{~~~~} c@{~~}c @{~~~~} c@{~~}c @{~~~~} c @{~~~~} c @{~~~~} c@{~~}}
        \toprule
        \multicolumn{3}{r}{\textbf{Task-specific architecture}} & \multicolumn{3}{c}{\hspace*{-2mm}ContextNet}& \multicolumn{3}{c}{\hspace*{-2mm}LSTM} & \multicolumn{2}{c}{\hspace*{-2mm}LSTM}    & \multicolumn{2}{c}{\hspace*{-2mm}LSTM}  & LSTM    & ECAPA & ECAPA  \\
        &&& \multicolumn{6}{c}{\hspace*{-2mm}\textbf{LibriSpeech train-clean-100}} & \multicolumn{2}{c}{\hspace*{-2mm}\textbf{Welsh 15.8h}} & \multicolumn{2}{c}{\hspace*{-2mm}\textbf{Basque 11h}} & \textbf{IC} & \textbf{ER} & \textbf{ASV} \\
        && & \multicolumn{6}{c}{\hspace*{-2mm}\textbf{WER $\downarrow$}} & \multicolumn{2}{c}{\hspace*{-2mm}\textbf{WER $\downarrow$}} & \multicolumn{2}{c}{\hspace*{-2mm}\textbf{WER $\downarrow$}} & \multicolumn{2}{c}{\hspace*{-2mm}\textbf{Acc. $\uparrow$}} & \textbf{EER $\downarrow$} \\
        \textbf{Context encoder}  & \textbf{Size} & \textbf{Trained on}
            & d.c.  & t.c.  & t.o & d.c.  & t.c.  & t.o & dev   & test  & dev   & test  & test    & test & test \\
        \cmidrule(r){1-3}\cmidrule(r){4-6}\cmidrule(r){7-9}\cmidrule(r){10-11}\cmidrule(r){12-13}\cmidrule(r){14-14} \cmidrule(r){15-15}\cmidrule(r){16-16}
        MHSA & 165M & \multirow{2}{*}{LL 4.3k h}  & 7.6 & 7.8  & 21.4  & 6.6 & 6.9  & 21.3  & 48.8  & 50.8  & 43.6   & 44.0   & 78.1 & \textbf{64.8} & 2.6  \\
        SummaryMixing & 155M &   & \textbf{7.3}  & \textbf{7.4} & \textbf{17.3} & \textbf{6.2}  & \textbf{6.6} & \textbf{17.2} & \textbf{45.7}  & \textbf{48.3}  & \textbf{41.4}  & \textbf{42.1}  & \textbf{80.5}   & 64.3 & \textbf{2.4} \\
        \cmidrule(r){1-3}\cmidrule(r){4-6}\cmidrule(r){7-9}\cmidrule(r){10-11}\cmidrule(r){12-13}\cmidrule(r){14-14} \cmidrule(r){15-15}\cmidrule(r){16-16}
        w2v2 base \cite{zaiem2023speech} & \phantom{0}95M & LS 960 h & ---   & ---  & --- & ---   & 6.2  & 14.9 & ---    & 54.5  & ---    & 51.3  & 77.7    & 73.2  & 2.8 \\
        w2v2 large \cite{zaiem2023speech} & 317M & LV 60k h & ---   & ---  & ---    & ---   & 3.7  & \phantom{0}9.3    & ---    & 45.4  & ---    & 38.0   & 79.0    & 68.4 & 3.2  \\
        HuBERT large \cite{zaiem2023speech} & 317M & LL 60k h & ---   & ---  & ---    & ---   & 3.6  & \phantom{0}8.1    & ---    & 51.2  & ---    & 46.2   & 80.1    & 71.6 & 3.8  \\
        data2vec large \cite{zaiem2023speech} & 314M & LL 60k h & ---   & ---  & ---    & ---   & 3.1  & \phantom{0}6.5    & ---    & 44.3  & ---    & 38.2   & 79.9    & 71.3 & 2.7  \\
        \bottomrule
    \end{tabular}
\end{table*}

\subsection{Downstream tasks}
\noindent\textbf{MP3S benchmark.} We test the downstream tasks performance of the w2v2 models through SpeechBrain MP3S benchmark \cite{zaiem2023speech}. 
Similar to other widely used SSL benchmarks such as SUPERB \cite{yang2021superb},
the pre-trained SSL models are frozen.
The weighted sum of the input to the context encoder and the hidden representations from all the context encoder layers, 
with respect to a set of trainable weights, 
is used as the speech representation for the downstream models.
However, differently from SUPERB, MP3S does not limit each task to a single downstream model.
The tasks in MP3S include automatic speech recognition (ASR), intent classification (IC), emotion recognition (ER), and automatic speaker verification (ASV). \\

\noindent\textbf{Experimental setups.} For ASR, following MP3S, we considered three datasets: LibriSpeech \cite{panayotov2015librispeech} train-clean-100 split for English ASR, CommonVoice 11.0 \cite{ardila2020common} Welsh (Cymraeg) and Basque (Euskera) datasets for low-resource ASR. 
The Welsh and Basque dataset contain 15.8 and 11 hours of training data, respectively.
We use the long short-term memory (LSTM)  \cite{hochreiter1997long} or ContextNet \cite{han2020contextnet} with Connectionist Temporal Classification (CTC) \cite{graves2006connectionist} objective function as the downstream model for LibriSpeech, 
and LSTM with CTC for Welsh and Basque.
We do not use the Buckeye corpus \cite{pitt2005buckeye} as in MP3S due to licence issues. 
For IC, we use the SLURP dataset \cite{bastianelli2020slurp} with a LSTM downstream model. 
For ER, we use the IEMOCAP dataset \cite{busso2008iemocap} with an ECAPA \cite{desplanques2020ecapa} downstream model. 
For ASV, we use the VoxCeleb1 corpus \cite{nagrani2017voxceleb} with an ECAPA downstream model. All the model and training configurations follow \cite{zaiem2023speech}. \\

\noindent\textbf{Experimental results.} Table~\ref{tab:res_asr} shows the performance of downstream tasks. 
For ASR, our SummaryMixing w2v2 model outperforms the MHSA w2v2 model for English LibriSpeech dataset with 100h training data.
For low-resource ASR with limited training data, SummaryMixing w2v2 surpasses MHSA w2v2 for both of Welsh 15.8h and Basque 11h. 
For all ASR tasks, on average, SummaryMixing reduces the WERs by 7.8\%, relatively. 
The w2v2 base model outperforms trained w2v2 Conformer models on LibriSpeech, 
but this is due to the unfair advantage that w2v2 base is also pre-trained with LibriSpeech.
For the low-resource ASR where the ASR datasets are strictly different from the pre-training datasets, SummaryMixing w2v2 outperforms the w2v2 base model by a large margin --
14.7\% relative WERs reduction on average.
More impressively, for low-resource ASR, our SummaryMixing w2v2 surpasses the HuBERT large model with a 7.3\% average relative WERs reduction. 
These results demonstrate the effectiveness of the SummaryMixing SSL model for ASR under difference scenarios.

Not included in the table is an intriguing initial result when the w2v2 model itself is finetuned as well, on LibriSpeech train-clean-100, using a two-layered neural network on top of it with the CTC objective function.
In this experiment, with MHSA the LibriSpeech dev-clean WER is 5.1, which surprisingly is better than SummaryMixing, at 5.4 (w2v2 base gives a 6.1 WER \cite{baevski2020wav2vec}). 
This is unexpected since when trained in a fully supervised fashion, SummaryMixing outperforms MHSA for a variety of ASR tasks \cite{parcollet2024summarymixing},
and the frozen SummaryMixing w2v2 also gives better downstream ASR results as shown in Table~\ref{tab:res_asr}.
We leave the investigation of this unexpected result as future work.

Table~\ref{tab:res_asr} also shows SummaryMixing w2v2 outperforms MHSA w2v2 for intent classification (IC) and automatic speaker verification (ASV), with a 2.0\% and 7.7\% relative performance gain, respectively. 
It is worth noting that for IC and ASV, our SummaryMixing w2v2 model surpasses w2v2 large, Hubert large, and data2vec large, even though these large models have about $2\times$ more parameters and are pre-trained with $15\times$ more data.
For emotion recognition (ER), SummaryMixing is slightly behind MHSA (0.7\% relative). 
However, for ER, the w2v2 large model is also 6.5\% relatively worse than the w2v2 small model.
This indicates it may be difficult to produce a universal optimal w2v2 model for all tasks.
Therefore, in summary, SummaryMixing gives better or equivalent results for the downstream ASR, IC, ER and ASV tasks compared to MHSA.

\begin{figure}[t!]
    \hspace*{-2mm}
    \includegraphics{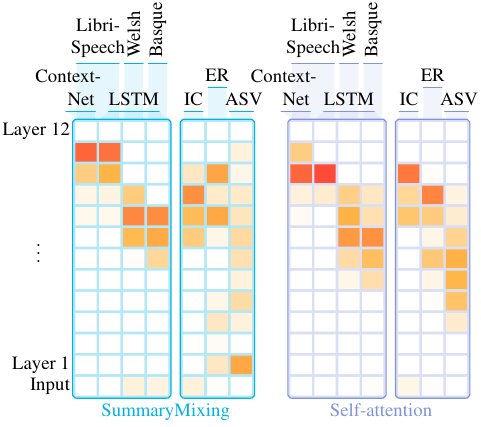}
    \caption{The learned weights of the SummaryMixing (left) and MHSA (right) Context encoder hidden representations  for downstream tasks. The weights of each column sum to one. Deeper colors indicate larger weights.}\label{fig:layer_weights}
\end{figure}

Figure~\ref{fig:layer_weights} shows the learned weights for each context encoder layer for all the downstream tasks. 
For ASR, IC, and ER, both the SummaryMixing and MHSA models have most weights assigned to the higher layers,
which is consistent with previous findings that the higher layers of trained w2v2 models tend to capture linguistic features such as phonetic information and word meaning \cite{pasad2021layer}.
For ASV, the downstream model tend to use low-level features, which is also consistent with the literature \cite{pasad2021layer}.
The distribution of the learned weights and the numerical experimental results indicate that SSL pre-training with the linear model SummaryMixing well captures different levels of representations, i.e., content (ASR), semantic (IC), paralinguistic (ER) and speaker (ASV) features.


\section{Conclusion}
In this paper, we investigate a linear-complexity model, a Conformer with SummaryMixing, as the context encoder for wav2vec 2.0 model. 
Compared to self-attention based Conformer context encoder, SummaryMixing improves the pre-training speed by 18\% and reduces the peak VRAM by 23\%. 
Also, when used as a feature extractor, with SummaryMixing the model yields better or the same level of performance for downstream speech processing tasks compared to wav2vec 2.0 with self-attention.
Future works include building other SSL models using SummaryMixing in a Conformer context encoder, as well as exploring full fine-tuning for SummaryMixing SSL models.

\bibliographystyle{IEEEtran}

\apptocmd{\thebibliography}{\fontsize{8}{8.5}\selectfont}{}{}

\bibliography{mybib}

\end{document}